\documentclass[sigconf, screen]{acmart} 
\AtBeginDocument{%
  }

\usepackage{soul}
\usepackage{url}
\usepackage[utf8]{inputenc}
\usepackage{graphicx}
\usepackage{amsmath}
\usepackage{amsthm}
\usepackage{booktabs}
\usepackage{algorithm}
\usepackage{algorithmic}
\usepackage[switch]{lineno}
\usepackage{balance}
\usepackage{colortbl}  
\usepackage{enumitem}
\usepackage{tikz}
\usepackage{multirow}
\usepackage{xspace}
\usepackage{tablefootnote}
\usepackage{booktabs}
\usepackage{subcaption}
\usepackage{multicol}
\usepackage{natbib}
\usepackage{pifont}  
\usepackage{cleveref}

\newcommand{\gain}[1]{\textcolor{green}{#1}}

\newcommand{\model}{LPD\xspace}
\newcommand{\dcLoss}{de-correlation loss\xspace}
\newcommand{\DcLoss}{De-correlation loss\xspace}
\newcommand{\efLoss}{\texttt{EF-MTRL}\xspace}
\def\ie{\textit{i.e.}~}
\def\wrt{\textit{w.r.t.}~}

\crefname{table}{Tab.}{Tables}
\crefname{figure}{Fig.}{Figures}
\crefname{section}{Sec.}{Sections}
\crefname{equation}{Eq.}{Equations}

\usepackage{xcolor}
\definecolor{iccvblue}{rgb}{0.21,0.49,0.74}
\definecolor{linkred}{rgb}{1.0,0,0}
\definecolor{refergreen}{rgb}{0.44,0.67,0.28}

\setcopyright{acmlicensed}
\copyrightyear{2025}
\acmYear{2025}
\acmBooktitle{Proceedings of the 33rd ACM International Conference on Multimedia (MM '25), October 27--31, 2025, Dublin, Ireland}
\acmConference[MM '25]{Proceedings of the 33rd ACM International Conference on Multimedia}{October 27--31, 2025}{Dublin, Ireland.}
\acmDOI{10.1145/3746027.3755476}
\acmISBN{979-8-4007-2035-2/2025/10}

\begin{document}
\hypersetup{
  linkcolor=iccvblue,      
  citecolor=refergreen,      
  urlcolor=iccvblue         
}

\title{Learning Partially-Decorrelated Common Spaces for Ad-hoc Video Search}

\author{Fan Hu}
\affiliation{
  \institution{Renmin University of China}
  \city{Beijing}
  \country{China}}
\email{hufan_hf@ruc.edu.cn}

\author{Zijie Xin}
\affiliation{
  \institution{Renmin University of China}
  \city{Beijing}
  \country{China}}
\email{xinzijie@ruc.edu.cn}

\author{Xirong Li}
\authornote{Corresponding author.}
\affiliation{
  \institution{Renmin University of China}
  \city{Beijing}
  \country{China}}
\email{xirong@ruc.edu.cn}

\renewcommand{\shortauthors}{Fan Hu, Zijie Xin, \& Xirong Li}

\begin{abstract}
Ad-hoc Video Search (AVS) involves using a textual query to search for \textit{multiple} relevant videos in a large collection of unlabeled short videos.
The main challenge of AVS is the visual diversity of relevant videos.
A simple query such as ``Find shots of a man and a woman dancing together indoors'' can span a multitude of environments, from brightly lit halls and shadowy bars to dance scenes in black-and-white animations. It is therefore essential to retrieve relevant videos as comprehensively as possible.
Current solutions for the AVS task primarily fuse multiple features into one or more common spaces, yet overlook the need for diverse spaces.
To fully exploit the expressive capability of individual features, we propose \model, short for \textbf{L}earning \textbf{P}artially \textbf{D}ecorrelated common spaces. 
\model incorporates two key innovations: feature-specific common space construction and the \dcLoss. 
Specifically, \model learns a separate common space for each video and text feature, and employs \dcLoss to diversify the ordering of negative samples across different spaces.
To enhance the consistency of multi-space convergence, we designed an entropy-based fair multi-space triplet ranking loss.
Extensive experiments on the TRECVID AVS benchmarks (2016-2023) justify the effectiveness of \model.
Moreover, diversity visualizations of \model's spaces highlight its ability to enhance result diversity.
\end{abstract}



\maketitle

\section{Introduction}

\begin{figure*}[t]
    \centering
    \begin{subfigure}{0.95\textwidth}
        \centering
        \includegraphics[width=\linewidth]{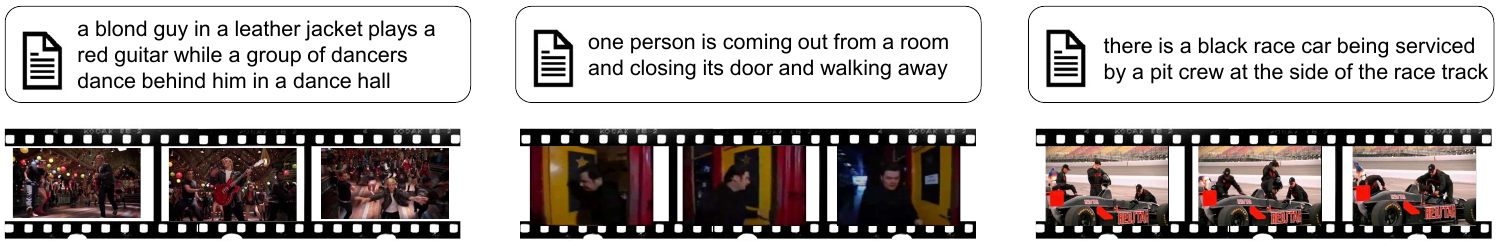}
        \caption{Text-to-video matching. Data: MSR-VTT \cite{msr-vtt}}
        \label{fig:t2v_task_visualization}
    \end{subfigure}
    
    \begin{subfigure}{0.95\textwidth}
        \centering
        \includegraphics[width=\linewidth]{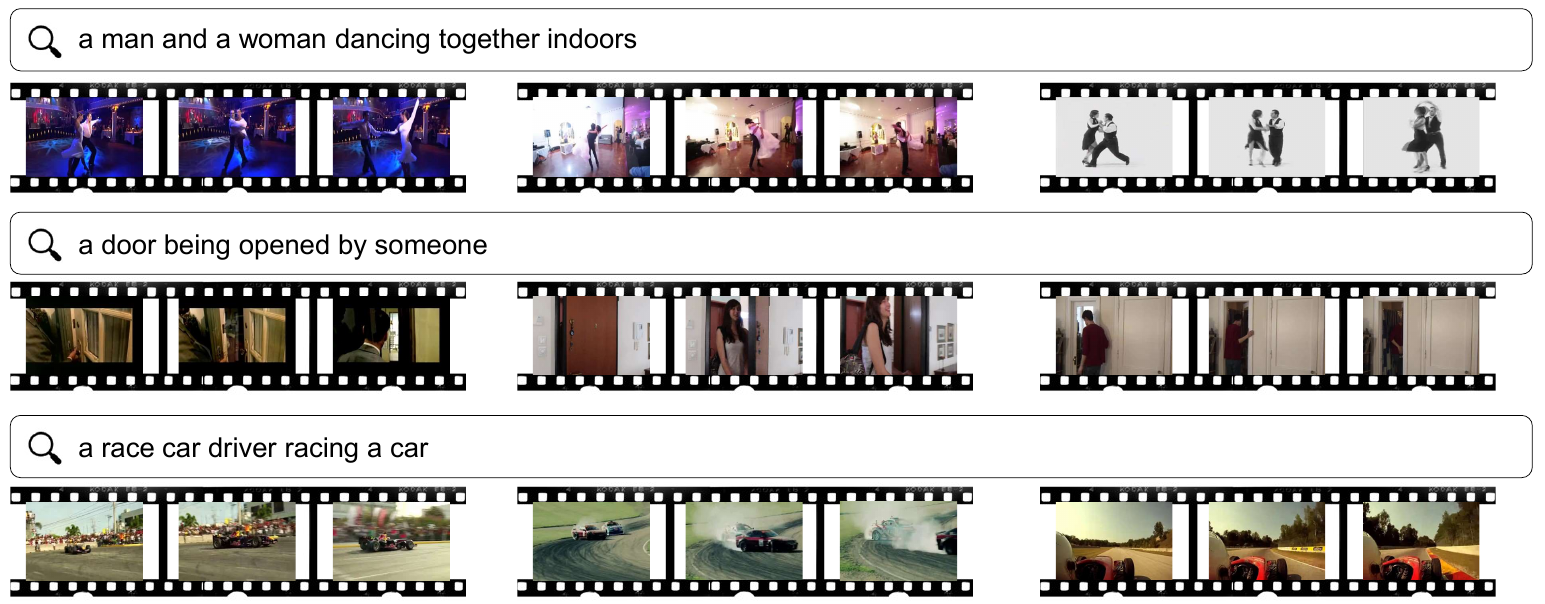}
        \caption{Ad-hoc video search. Data: TRECVID \cite{Trecvid2019}}
        \label{fig:avs_task_visualization}
    \end{subfigure}

    \caption{\textbf{Illustrating two types of text-to-video retrieval tasks}: (a) text-to-video matching (TVM) and (b) ad-hoc video search (AVS). Compared to TVM which is to find a single match within thousands of videos, AVS aims to retrieve \emph{multiple} relevant videos from a \emph{million}-scale collection. This paper is targeted at the latter.}
    \Description{xxx}
    \label{Figure:two_task_visualize}
\end{figure*}

Ad-hoc video search (AVS), or zero-example video retrieval, involves identifying multiple relevant videos through text-based queries without visual references. Since 2016, the annual TRECVID (TV) evaluation has been a benchmark for measuring progress on the AVS task \cite{Trecvid2016}. Each team is asked to develop a video retrieval system that retrieves the top 1,000 items for each test query from a large collection of unlabeled short videos. The organizers then annotate the items submitted by all teams and announce the scores and rankings.
AVS is different from text-to-video matching (retrieval) task, which seeks to identify the \textit{most} semantically relevant video clip from a candidate pool referring to the textual query \cite{jiang2023dual, xclip, clip4clip, ts2net}. AVS aims to retrieve all videos related to the textual query, which may encompass a variety of scenarios \cite{TRECVid2017,Trecvid2018,Trecvid2019,Trecvid2020,Trecvid2021,Trecvid2022,Trecvid2023}. The visualization examples between these two tasks are illustrated in \cref{Figure:two_task_visualize}.

The AVS task is particularly challenging for two key reasons. Firstly, it requires searching from an extremely large collection of unlabeled short videos. The size of the test collection for AVS TV 2016--2018 (IACC.3 dataset \cite{Trecvid2016}) is 336k, which increases to one million for TV 2019--2021 (V3C1 dataset \cite{V3C1}) and 1.5 million for TV 2022--2023 (V3C2 dataset \cite{Trecvid2022}). The second challenge is that AVS task not only aims to accurately locate videos that match the query, but also comprehensively retrieve relevant videos across all possible scenes.
The complexity is heightened by the potential variability in the relevant video content, including scenes set in broad daylight, during the night, or even within animated environments. As shown in \cref{Figure:two_task_visualize}, a simple query like ``\emph{Find shots of a man and a woman dancing together indoors}'' could span a multitude of environments, from brightly lit halls and shadowy bars to dance scenes within black and white animations, which highlight the assessment challenges outlined in annual TRECVID reports  \cite{Trecvid2016,TRECVid2017,Trecvid2018,Trecvid2019,Trecvid2020,Trecvid2021,Trecvid2022,Trecvid2023}.

Currently, the end-to-end task-specific models based on CLIP \cite{clip} are popular in text-to-video matching task \cite{clip4clip, xclip, clip_vip_XueS0FSLL23, wu2023cap4video, jin2023diffusionret}. However, given the considerable challenges posed by the large size of video pools and the diversity of video scenes, the winning models of AVS task are primarily based on deep learning models that extract diverse (off-the-shell) textual or video features, followed by cross-modal training and matching \cite{tv19-rucmm, tv20-lirenmin,tv20-wuvireo, tv21-wu2021vireo, tv21-Kuaishou, tv21-lirenmin, tv22-Waseda, tv22-lirenmin, tv22-ITI_CERTH, tv23-Waseda, tv23-lirenmin}. 
We categorize them into two groups based on the number of common spaces:

The first category is learning one common space with multiple feature fusion, such as DE \cite{dualencoding, tv18-rucmm, tv19-rucmm, tv20-aim3, tv21-wu2021vireo} and W2VV++ \cite{w2vv++, tv18-rucmm, tv19-rucmm}, fusing multiple features/models to create a stronger composite feature.  However, learning within a singular common space could present challenges in achieving diversity and comprehensive recall.
The second category is centered around learning multiple common spaces with different features, such as creating a concept space and an embedding space \cite{wu2020interpretable_dual_task, tv20-wuvireo, tv21-wu2021vireo}, multi-level video-text matching \cite{chen2020hgr, tv20-aim3}, aligning each text encoder with a unique common space \cite{sea, tv20-lirenmin, tv21-lirenmin}, and generating multiple spaces that enable cross-matching between various text and video features \cite{t_multi_v_GalanopoulosM22, tv22-ITI_CERTH}.
Nonetheless, these methods lack a specific design for enhancing the diversity of features across different spaces. Moreover, we believe that each feature, whether text or video, needs to learn its dominant space. And there is also a need for tailored training strategies that not only ensure accuracy but also enhance the diversity of retrieval results. 

In this paper, we introduce a new network architecture, \model (\textbf{L}earning \textbf{P}artially-\textbf{D}ecorrelated Common Spaces for Ad-hoc Video Search). The overall framework is depicted in \cref{fig:overall_framework}. To fully exploit the expressive capability of each feature within its corresponding space, we design the architecture so that one end of each space is connected to the corresponding feature, while the other end integrates a weighted fusion of features from the corresponding modality. To enhance the diversity of retrieval results across multiple spaces while ensuring relevance, we introduce a \dcLoss that imposes constraints on the ordering of negative samples differently in each space. To enhance the consistency of multi-space convergence, we designed the entropy-based fair multi-space triplet ranking loss.
By integrating the relevance-enhancing triplet ranking loss with the diversity-promoting \dcLoss, \model not only maintains the relevance of the retrieval outcomes but also boosts their diversity across various spaces.
What's more, \dcLoss is model-agnostic and can be applied to other multi-space methods as well.
In summary, our main contributions are as follows:
\begin{itemize}
\item To improve result diversity in the AVS, we propose a novel model \model, which learns multiple feature-specific common spaces, exploiting the expressive capability of each feature.

\item We design the \dcLoss to enhance diversity and improve the multi-space triplet ranking loss by enforcing fair space convergence.

\item Experiments on the long-standing TRECVID AVS benchmark series (2016--2023) verify the effectiveness of \model. Source code is available at \url{https://github.com/xxayt/LPD}.
\end{itemize}

\section{Related Work}

\subsection{Multi-feature Video Retrieval}
Existing research on multi-feature video retrieval has developed various common-space construction styles.
Single-space methods typically fuse multiple features into a robust composite feature for matching within a common space. For example, W2VV++ \cite{w2vv++} and DE \cite{dualencoding} integrate multiple features through vector concatenation, while CE \cite{ce_bmvc_2019} employs a collaborative experts model to aggregate features. 
Alternatively, some methods define several fixed common spaces that are not feature-specific. DualTask \cite{wu2020interpretable_dual_task} constructs an embedding space and a concept space. LAFF \cite{laff} uses a multi-head approach to establish multiple parallel common spaces. TMVM \cite{lin2022TMVM} constructs a multi-embedding space by learning diverse visual prototypes for each video.
Further, some methods construct common spaces tailored to specific features. MMT \cite{2020mmt} used a multi-modal transformer to fuse multiple video features and create video-feature specific spaces. SEA \cite{sea} aligned each text feature with a unique common space.
T$\times$V \cite{t_multi_v_GalanopoulosM22} performed cross-matching for each text-video feature pair, resulting in a quadratic number of common spaces.
Improved-ITV \cite{improved_itv_icmr2024} employed generative captions for multi-stage fusion but ignores feature complementarity and space diversity.
Our method differs by directly promoting diversity through space-wise regularization during training.
In this paper, we propose \model, which learns feature-specific spaces for all video and text features, coupled with a \dcLoss designed to enhance the diversity across different spaces.
The number of common spaces in \model is equal to the number of features, making it more efficient than T$\times$V.

\subsection{Ad-hoc Video Search}
Ad-hoc video search (AVS), evaluated annually in TRECVID, has been studied since 2016 \cite{Trecvid2016}.
In this task, given a textual topic description, the search system formulates a query and returns a ranked list of relevant video shots.
Early efforts matched concept representations extracted from both videos and queries using predefined concept sets \cite{Snoek:concept_video_retrieval, Naphade:LSCOM, Jiang:semantic_concept, Snoek:challenge, tv16-NII2016, tv18-Waseda2018, tv18-Informedia2018}. However, this required substantial manual effort for concept design and annotation \cite{anh:KIS, tv19-waseda}.

With advancements in deep neural networks and the availability of video captioning datasets, recent progress in ad-hoc video search typically employs visual-text embedding models \cite{videostory,vse,w2vv++,dualencoding,wu2020interpretable_dual_task, sea,chen2020hgr, laff,t_multi_v_GalanopoulosM22}. Due to the need for comprehensive retrieval across diverse scenes in the AVS, teams typically employ deep models to extract multiple features and map them into one or more common spaces to calculate cross-modal similarity \cite{tv18-rucmm, tv19-rucmm, tv20-aim3, tv20-lirenmin, tv20-wuvireo, tv21-wu2021vireo, tv21-Kuaishou, tv21-lirenmin, tv22-Waseda,tv22-lirenmin, tv23-Waseda, tv23-lirenmin}.
Some methods learn a single common space by fusing multiple features and enhance performance with additional training data and advanced features \cite{tv18-rucmm, tv19-rucmm, tv20-aim3, tv21-wu2021vireo}. Others design interpretable multi-space architectures for matching \cite{tv20-wuvireo, tv21-wu2021vireo}, or construct multiple text-specific spaces and perform cross-matching \cite{tv20-lirenmin, tv21-Kuaishou, tv21-lirenmin, tv22-ITI_CERTH}. More recently, multi-model late fusion has also been explored \cite{tv22-Waseda,tv22-lirenmin,tv22-ITI_CERTH,tv23-whu, tv23-Waseda, tv23-lirenmin}. 
Beyond space construction, relevance-aware mining \cite{falcon2024improving} and support-set training \cite{support_iclr2021} have been proposed to better leverage unlabeled positives. Separately, general-purpose diversification methods such as MMR \cite{mmr_sigir1998} and xQuAD \cite{xquad_2012} improve diversity in retrieval but may sacrifice relevance. 
In this paper, we construct diverse common spaces for multiple features and achieve a better trade-off without post-processing, making our approach more practical for AVS.

\section{Proposed LPD Method}

\begin{figure*}[!htb]
\centering
    \subfloat[Construction of multiple feature-specific common spaces 
    \label{fig:overall_framework}]{\includegraphics[width=0.98\columnwidth]{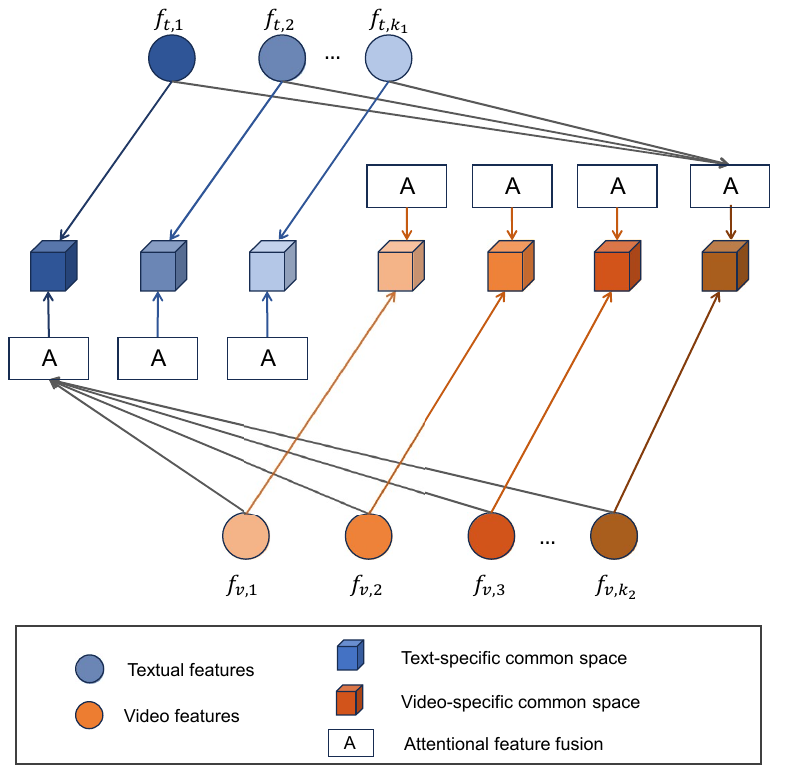}} 
    \hfill
    \subfloat[Calculating  proposed \dcLoss for spaces $m$ and $n$
    \label{fig:de_correlation_loss}]
    {\includegraphics[width=0.98\columnwidth]{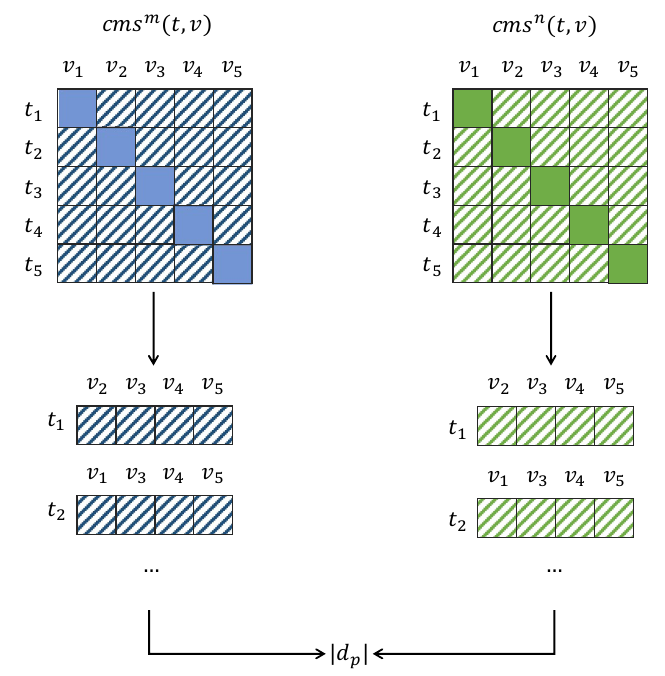}}  
    \caption{ \textbf{Proposed \model}.  
    Each common space for video-text matching is feature-specific, with one end connected to its corresponding feature and the other end as a weighted fusion of features from the other modality. The feature transform layers associated with the same feature (indicated by the arrows originating from the feature in the figure) share parameters.
    For the \dcLoss in a mini-batch with three video-text pairs, the blue and green squares represent the text-video similarity in spaces $m$ and $n$, respectively. Different rows correspond to different text queries, and different columns represent different videos. The goal is to improve the diversity of retrieval outcomes across various spaces by lowering Pearson correlation coefficient for negative videos across these spaces. 
    }
    \label{Figure:model_set}
\end{figure*}

\subsection{Overall Framework}

We formalize an ad-hoc video search process as follows. We denote a specific video clip as $v$ and a large collection of $n$ \emph{unlabeled} video clips as $\mathcal{V}=\{v_1,\ldots,v_n\}$. For an ad-hoc query in the form of a sentence $t$, let $\text{cms}(t,v)$ be a \textbf{c}ross-\textbf{m}odal \textbf{s}imilarity function that measures the semantic relevance between the query and a specific video. Accordingly, the search process boils down to sorting $\mathcal{V}$ in descending order in terms of $\text{cms}(t,v)$ and returning the top-ranked items for the given query. The computation of $\text{cms}(t,v)$ requires proper embeddings of both $t$ and $v$ into a common cross-modal space. 

For features, suppose all textual queries are represented by a set of $k_1$ sentence-level features $\{f_{t,1}, \ldots, f_{t,k_1}\}$, and the videos are represented by a set of $k_2$ video-level features, $\{f_{v,1}, \ldots, f_{v,k_2}\}$. 

We need to establish a common space for each text feature and video feature, wherein we compute the similarity. As the features are obtained by distinct extractors and thus incompatible, we shall use a feature transformation layer to rectify the diverse features to the same length embeddings.
Then, we could get  sentence-level embeddings $\{e_{t,1}, \ldots, e_{t,k_1}\}$, and video-level embeddings, $\{e_{v,1}, \ldots, e_{v,k_2}\}$. 
Since each embedding independently dominates a common space, the cross-modal similarity $\text{cms}(t,v)$ is the average similarity across all common spaces, \ie

\begin{equation}
\left\{ 
\begin{array}{ll}
\begin{aligned}
cms^i(t,v) &= cosine [e_{t,i},\, \text{Fusion}_{v,i}(e_{v,1},\ldots,e_{v, k_2})], \\
cms^j(t,v) &= cosine [\text{Fusion}_{t,j}(e_{t,1},\ldots,e_{t, k_1}),\, e_{v,j}],\\
cms(t,v) &= \dfrac{1}{k_1+k_2} \left[\sum_{i=1}^{k_1} cms^i(t,v) +  \sum_{j=1}^{k_2} cms^j(t,v)\right].
\end{aligned}
\end{array} \right.
\label{eq:feature_space}
\end{equation}

Where $\text{cms}^i(t,v)$ and $\text{cms}^j(t,v)$ represent the similarities of the common spaces dominated by the \( i^{th} \) text feature and the \( j^{th} \) video feature, respectively. 
Thus, for each common space, one end is the embedding obtained from an individual feature, and the other end is the fusion of embeddings from another modality. Different spaces aim to mine the information from various features, enhancing the diversity of retrieval and recalling more relevant videos.

Next, we describe in detail of feature transform and fusion, followed by the proposed \dcLoss and fair multi-space triplet ranking loss.

\subsection{Feature Transform and Fusion} \label{ssec:Feature_transfrom}

Without loss of generality, we are provided with a diverse set of $k$ different features $\{f_1, \ldots, f_k\}$, sized as $d_1, \ldots, d_k$, respectively. As the features are obtained by distinct extractors and thus incompatible, we shall use a feature transformation layer to rectify the diverse features to be of the same length. To convert the $i$-th feature to a new $d$-dimensional feature embedding, we use 
\begin{equation} \label{eq:transform}
e_i = \phi \left(\text{Linear}_{d_i \times d} (f_i)\right),
\end{equation}
where $\phi$ is a nonlinear activation function. Following previous work \cite{w2vv++,sea,laff}, we use $\tanh$ as $\phi$. 

Regarding the fusion method, as indicated in \cref{eq:feature_space}, each feature-specific common space is designed to match a specific feature with a dynamic weighted fusion of features from the opposite modality. 
Consequently, we contemplate utilizing a weighted fusion to integrate the features ${e_1, \dots, e_k}$ from the other modality, where $k=k_1$ for text and $k=k_2$ for video
\begin{equation} \label{eq:combine_feats}
\text{Fusion}(e_1, \dots, e_k) = \sum_{i=1}^k a_i e_i,
\end{equation}
with weights $\{a_1, \ldots, a_k\}$ computed by a attention layer as:
\begin{equation}\label{eq:light-att}
\{a_1, \ldots, a_k\} = \mbox{softmax}\left(\text{Linear}_{d\times1}(\{e_1, \dots, e_k\})\right).
\end{equation}

\subsection{De-correlation Loss} \label{ssec:Dr_ranking}

To enhance the complementarity between the multiple common spaces while preserving their ability to retrieve positive samples, we propose an auxiliary loss that minimizes the correlation between the rankings of negative samples derived from the individual spaces.
Since our method uses multi-embedding spaces, each space produces a ranking, and the final retrieval result is obtained by score-based fusion of these rankings, in \cref{eq:feature_space}. By minimizing the correlation between individual rankings, we naturally promote diversity in the final retrieval result, thereby enhancing its diversity.

As shown on \cref{fig:de_correlation_loss}, for a specific space $m$, the $b \times b$ video-text similarity matrix is represented as $\text{cms}^m(t,v)$.
Accordingly, the $i$ th row $\text{cms}^m(t_i,v)$ contains one positive video-text sample pair similarity $\text{cms}^m(t_i,v^{+})$ and $b-1$ negative video-text sample pair similarities $\text{cms}^m(t_i,v^{-})$.

To apply the \dcLoss, for each text in different spaces, we apply a constraint to increase the dispersion of these rankings among different spaces, thereby enhancing the diversity of retrieval outcomes. For two specific spaces (indexed by $m$ and $n$) from the $(k_1+k_2)$ available spaces, the \dcLoss $\texttt{DcL}$ is defined as:
\begin{equation} \label{eq:dcl}
\texttt{DcL} = \frac{1}{b} \sum_{i=1}^{b}  \left|d_p\left(\text{cms}^m\left(t_i,v^{-}\right), \text{cms}^{n}\left(t_i,v^{-}\right)\right)\right|,
\end{equation}
where $d_p$ is the Pearson correlation coefficient, which is applied between the ordered sets of negative samples across different spaces for a given text. We aim to optimize the absolute value of $d_p$ to 0 because our goal is to enhance differences in retrieval outcomes across different spaces, rather than making the retrieval outcomes completely opposite with $d_p=-1$.

To facilitate parallel processing of \cref{eq:dcl}, we utilize a masking method for matrix operations on the video-text similarity matrices. Let $M$ be the similarity matrix for a specific space.
The $i$-th row $M_{i}$ stores similarity scores of text $t_i$ to all videos in the batch. The diagonal elements represent positive sample pairs, while the off-diagonal elements represent negative sample pairs. 
In order to mask out the positive pairs when computing the row-wise rank correlation coefficient, we define a binary matrix $Z$, where $Z_{i,j}=0$ for $i=j$ and $Z_{i,j}=1$ otherwise. 
Applying element-wise multiplication with $Z$ on the similarity matrices \wrt space $m$ and $n$, \texttt{DcL} can be rewritten as:
\begin{equation}
\texttt{DcL} = \frac{1}{b} \sum_{i=1}^{b} \left|d_p\left((M^m \odot Z)_i, (M^n \odot Z)_i\right)\right|.
\end{equation}
This allows for direct matrix operations for efficient computation of \texttt{DcL} and thus accelerates training.

\subsection{Fair Multi-space Triplet Ranking Loss} \label{ssec:multi_space_loss}
During each training iteration, we sample a mini-batch $\mathrm{B}$ containing $b$ video-text pairs $\{(v_i, t_i) \mid i=1, \ldots, b\}$ from the training dataset. This sampling strategy ensures that each video is irrelevant to other texts within the batch, and vice versa.

For a specific space $s$, we use the improved triplet ranking loss (ITRL) by  \citet{vsepp}. While originally proposed for image-text matching, ITRL is now found to be effective for text-video matching~\cite{mithun-icmr18,w2vv++,sea,laff,dualencoding}. Unlike the classical triplet ranking loss that selects negative training examples by random, ITRL considers the negative that violates the ranking constraint the most (within a mini-batch) and is thus deemed to be the most informative for improving the model being trained. Given a training sentence $t$ with $v^+$ as a video relevant \wrt $t$ and $v^-$ as irrelevant, we express the ITRL of a specific space $s$ as 
\begin{equation} \label{eq:triplet_ranking_loss}
\left\{ \begin{array}{ll}
 v^{-*} &= \mbox{argmax}_{v^- \in \mathrm{B}} \left(\text{cms}^s\left(t,v^-\right) - \text{cms}^s\left(t,v^+\right) \right), \\[1mm]
 \texttt{ITRL}^s &=\max\left(0, \alpha + \text{cms}^s\left(t,v^{-*}\right) -\text{cms}^s\left(t,v^+\right)\right), \end{array} \right.
\end{equation}
where $\alpha$ is a positive hyper-parameter concerning the margin.

However, applying ITRL to multiple spaces presents a challenge. Due to the diversity across different spaces, convergence rates often vary, and directly summing the ITRL from different spaces may lead to overfitting in some spaces. This phenomenon arises because different features, which dominate different spaces, exhibit considerable variations in dimensions, representational capabilities for different modal information, and update frequencies, leading to disparate convergence speeds among spaces. If a space has converged, subsequent training is likely to result in overfitting \cite{james2013introduction, goodfellow2016deep}.

Intuitively, spaces with slower convergence rates, wherein feature distribution updates less frequently, require more extensive training. Inspired by the entropy of feature \cite{ma2021towards} and One-shot Supernet \cite{wei2023automatic}, we propose a multi-space triplet ranking loss with fair space convergence. It dynamically selects different spaces for training based on the information entropy of features, promoting a more balanced and effective training process across all spaces and leading to more stable convergence.
Next, we detail the method for calculating space weights based on feature information entropy and the process for computing an improved multi-space loss.

\subsubsection{Feature Entropy based Common Space Weighting}

To compute the feature information entropy for space $s$, we collect all its features in the given batch and apply min-max normalization independently to each dimension. This yields a set of $N = b \times d$ values in the range $[0, 1]$. These values are then quantized into 100 bins, converting them into a 100-dimensional probability distribution $P_s$. The entropy $h_s$ is computed as:  
\begin{equation}
h_s = -\sum_{j=1}^{100} P_{s,j} \log(P_{s,j} + \epsilon),
\end{equation} 
where $\epsilon$ is a small positive constant ensuring numerical stability.   Given the $k_1+k_2$ distinct spaces, we obtain a space-wise entropy vector $H = \{h_1, h_2, \ldots, h_{k_1+k_2}\}$.

A space-wise weight vector, denoted as $W = \{w_1, w_2, \ldots, w_{k_1+k_2}\}$, is computed based on $H$ as:
\begin{equation} \label{eq:space-weights}
   W =  \mbox{softmax}(\tanh(H)).
\end{equation}

\subsubsection{Multi-space loss}
Once the space weight vector $W$ is dynamically given per iteration, each space is either included for or excluded from training, depending on whether its weight exceeds a given threshold (which is $\frac{1}{k_1+k_2}$). 
The final multi-space loss function, denoted as \efLoss (Entropy-based Fair Multi-space Triplet Ranking Loss) is computed as:
\begin{equation} \label{eq:ef-mtrl}
    \efLoss=\sum^{k_1+k_2}_{s=1} [w_s > \frac{1}{k_1+k_2}] \cdot \texttt{ITRL}^s.
\end{equation}

\subsection{Overall Loss}
The overall loss function is simply the sum of \texttt{DcL} and \efLoss.
By integrating these two individual losses, our model is motivated to not only maintain the relevance of the retrieval outcomes but also to boost their diversity across various spaces. This balanced approach tackles the dual challenge of ensuring both relevance and diversity in AVS, leading to a more effective and encompassing search performance.

\section{Evaluation}

\subsection{Experimental Setup}

\subsubsection{Test data}

We adopt the TRECVID evaluation \cite{Trecvid2016}, the \emph{de facto} international benchmark for AVS. 
Since the AVS 2024 data has not been released, we conduct experiments on the TV16-23 benchmark series, see \cref{table:test_dataset}.

\begin{table}[!tbh]
\centering \setlength{\tabcolsep}{10pt}
\caption{\textbf{Test sets used by the TRECVID (TV) AVS benchmark series}.  Frames are obtained by uniform sampling with a fixed time interval of 0.5 seconds.}

\renewcommand{\arraystretch}{1} 
\resizebox{0.9\linewidth}{!}{
\begin{tabular}{@{}lrrrr@{}}
\toprule
\textbf{Test set} & \multicolumn{1}{l}{\textbf{Video clips}} & \multicolumn{1}{r}{\textbf{Frames}} & \multicolumn{2}{r}{\textbf{Test queries}} \\
\midrule
\multirow{3}{*}{IACC.3 \cite{Trecvid2016}}  & \multirow{3}{*}{335,944} & \multirow{3}{*}{3,845,221} & TV16 & 30 \\
 & & & TV17 & 30 \\
 & & & TV18 & 30 \\ \midrule
\multirow{3}{*}{V3C1 \cite{Trecvid2019}} & \multirow{3}{*}{1,082,649} & \multirow{3}{*}{7,839,450} & TV19 & 30 \\
 & & & TV20 & 20 \\
 & & & TV21 & 20 \\ \midrule
\multirow{2}{*}{V3C2 \cite{Trecvid2022}}  & \multirow{2}{*}{1,425,443} & \multirow{2}{*}{6,082,291} & TV22 & 30 \\
 & & & TV23 & 20 \\ \bottomrule
\end{tabular}
}
\label{table:test_dataset}
\end{table}

\subsubsection{Development data} 

For training, we adopt the 9K subset of MSR-VTT \cite{yu2018joint}. Following \cite{tv18-rucmm,tv19-rucmm}, we adopt the TV16 video-text matching test set of 200 video-text pairs (TV16-VTT) as our validation set for model selection.

\subsubsection{Performance metric}

We follow the TRECVID protocol, reporting inferred Average Precision (infAP) \cite{Trecvid2016}.

\subsubsection{Test of significance}
We perform a randomization test to check if the performance difference between two video retrieval systems is statistically significant \cite{tmm2018-cross-media}.

\subsubsection{Implementations}
The networks have an output dimension of $d=512$. The margin $\alpha$ in \cref{eq:triplet_ranking_loss} is empirically set to 0.2.
Training is performed using SGD with a mini-batch size of 128, utilizing RMSProp as the optimizer. The initial learning rate is $10^{-4}$, reduced by a factor of 0.99 after each epoch.  We adopt mean Average Precision (mAP) for model selection: early stop occurs once there is no improvement in mAP on the validation set for ten consecutive epochs. 
We use six video features and three text features (\cref{tab:feature_intro}). 
All experiments are conducted with four 3090 GPUs.

\subsection{Comparison with Baselines} \label{compare_with_baaseline}

\newcommand{\tabincell}[2]{\begin{tabular}{@{}#1@{}}#2\end{tabular}}
\begin{table}[t!]
\setlength{\abovecaptionskip}{1pt}

\caption{\textbf{Video/text features used in our evaluation}. Video-level features are obtained by mean pooling over frames or segments unless otherwise stated.}
\label{tab:feature_intro}
\resizebox{\columnwidth}{!}{
\scriptsize
\centering
\begin{tabular}{@{}lrl@{}}
    \toprule
    \textbf{Feature} &  \tabincell{l}{\textbf{Dim.}}  & \textbf{Short description }       \\ 
    \noalign{\smallskip}
    \hline
    \noalign{\smallskip}
    \multicolumn{3}{@{}l}{\textbf{\textit{Video features:}}} \\ 
    \textit{wsl}  & 2,048 & \tabincell{l}{ResNeXt-101  pre-trained on 940  million  public \\ images, fine-tuned on ImageNet1k \cite{wsl2018}.}  \\
    \cmidrule{3-3}
    \textit{clip-b32}  & 512  & \tabincell{l}{CLIP (ViT-B/32)   pre-trained on web-scale image-text \\ pairs  by contrastive learning \cite{clip}.} \\  
    \cmidrule{3-3}
    \textit{clip-l14}  & 768  & \tabincell{l}{CLIP (ViT-L/14), with a  larger vision transformer.} \\ 
    \cmidrule{3-3}
    \textit{ircsn} & 2,048 & \tabincell{l}{irCSN-152  trained by weakly supervised learning on \\ IG-65M \cite{ircsn}.}  
    \\  
    \cmidrule{3-3}
    \textit{beit} & 2,048 & \tabincell{l}{BEiT pre-trained on full ImageNet and \\ fine-tune on 1k-class ImageNet  \cite{beit}.}   \\  
    \cmidrule{3-3}
    \textit{blip} & 256 & \tabincell{l}{BLIP(ViT-B) pre-trained on 129M \\ image-text pairs \cite{blip2022}.}   \\  
    \midrule
    \multicolumn{3}{@{}l}{\textbf{\textit{Text features:}}} \\ 
    \textit{blip}    & 256  & Text transformer from  BLIP(ViT-B). \\
    \textit{clip-b32}    & 512  & Text transformer from CLIP(ViT-B/32). \\
    \textit{clip-l14}    & 768  & Text transformer from  CLIP(ViT-L/14). \\ \bottomrule 
    \end{tabular}
}
\end{table}

\begin{table*}[t]
\centering \setlength{\tabcolsep}{8pt}

\caption{\textbf{Comparison with baselines}. All the multi-feature methods (Group III and IV) use the same set of video / text features (\cref{tab:feature_intro}). \model outperforms other methods significantly (p$<$0.05).}
\label{table:compare_baseline}

\renewcommand{\arraystretch}{1.1} 
\resizebox{0.85\linewidth}{!}{
\begin{tabular}{@{}p{2cm}rrrcrrrcrrr@{}}
\toprule
\multirow{2}{*}{\textbf{Method}} & \multicolumn{3}{c}{\textbf{IACC.3}} & & \multicolumn{3}{c}{\textbf{V3C1}} & & \multicolumn{2}{c}{\textbf{V3C2}} & \multicolumn{1}{r@{}}{\multirow{2}{*}{\textbf{MEAN}}} \\
\cmidrule{2-4} \cmidrule{6-8} \cmidrule{10-11}
& \multicolumn{1}{l}{\textit{TV16}} & \multicolumn{1}{l}{\textit{TV17}} & \multicolumn{1}{l}{\textit{TV18}} & & \multicolumn{1}{l}{\textit{TV19}} & \multicolumn{1}{l}{\textit{TV20}} & \multicolumn{1}{l}{\textit{TV21}} & & \multicolumn{1}{l}{\textit{TV22}} & \multicolumn{1}{l}{\textit{TV23}} & \multicolumn{1}{l}{} \\
\midrule
\multicolumn{8}{@{}l}{\textbf{\textit{Zero-shot retrieval}}:} \\ [2pt]
clip-b32 & 0.173 & 0.208 & 0.087 & & 0.136 & 0.161 & 0.194 & & 0.119 & 0.102 & 0.148 \\
clip-l14 & 0.201 & 0.232 & 0.103 & & 0.082 & 0.102 & 0.169 & & 0.118 & 0.116 & 0.140 \\
blip & 0.196 & 0.217 & 0.121 & & 0.159 & 0.203 & 0.223 & & 0.144 & 0.157 & 0.178  \\
\midrule

\multicolumn{8}{@{}l}{\textbf{\textit{CLIP-based end-to-end networks}}:} \\ [2pt]
CLIP4Clip \cite{clip4clip} & 0.196 & 0.228 & 0.108 & & 0.142 & 0.161 & 0.183 & & 0.127 & 0.139 & 0.161 \\
X-CLIP \cite{xclip} & 0.209 & 0.229 & 0.114 & & 0.150 & 0.184 & 0.195 & & 0.125 & 0.129 & 0.167 \\
TS2-Net \cite{ts2net} & 0.202 & 0.266 & 0.132 & & 0.120 & 0.153 & 0.188 & & 0.150 & 0.133 & 0.168 \\
CLIP-VIP \cite{clip_vip_XueS0FSLL23} & 0.184 & 0.212 & 0.109 & & 0.143 & 0.148 & 0.175 & & 0.109 & 0.088 & 0.146 \\
DGL \cite{yang2024dgl} & 0.194 & 0.224 & 0.121 & & 0.139 & 0.136 & 0.159 & & 0.110 & 0.111 & 0.149 \\
TeachCLIP \cite{2024TeachCLIP} & 0.199 & 0.248 & 0.129 & & 0.176 & 0.196 & 0.218 & & 0.166 & 0.164 & 0.187 \\
\midrule

\multicolumn{8}{@{}l}{\textbf{\textit{Multi-feature, single-space}}:} \\ [2pt]
W2VV++ \cite{w2vv++} & 0.215 & 0.301 & 0.138 & & 0.189 & 0.226 & 0.213 & & 0.177 & 0.186 & 0.206 \\
DE \cite{dualencoding} & 0.204 & 0.319 & 0.138 & & 0.189 & 0.235 & 0.225 & & 0.172 & 0.200 & 0.210 \\ 
\midrule

\multicolumn{8}{@{}l}{\textbf{\textit{Multi-feature, multi-space}}:} \\ [2pt]
DualTask \cite{wu2020interpretable_dual_task} & 0.209 & 0.332 & 0.144 & & 0.192 & 0.240 & 0.224 & & 0.178 & 0.203 & 0.215 \\
LAFF \cite{laff} & 0.222 & 0.287 & 0.142 & & 0.192 & 0.225 & 0.237 & & 0.175 & 0.183 & 0.208 \\
SEA \cite{sea} & 0.231 & 0.314 & 0.158 & & 0.199 & 0.251 & 0.230 & & 0.188 & 0.220 & 0.224 \\
T $\times$ V \cite{t_multi_v_GalanopoulosM22} & 0.240 & 0.313 & 0.159 & & 0.211 & 0.243 & 0.260 & & 0.198 & 0.200 & 0.228 \\
\rowcolor{green!8}\model & \textbf{0.245} & \textbf{0.327} & \textbf{0.169} & & \textbf{0.218} & \textbf{0.287} & \textbf{0.269} & & \textbf{0.221} & \textbf{0.221} & \textbf{0.245} \\
\bottomrule
\end{tabular}
}
\end{table*}

\subsubsection{Baselines} 
To ensure a fair and reproducible comparison, we select baselines with the following three criteria: 1) open source, 2) bi-encoder style for million-scale video retrieval as AVS, and 3) peer reviewed. As such, we obtain 15 baselines in four groups: \\
$\bullet$ Group I (zero-shot retrieval): Directly match the text embedding with the average pooled embedding of all frames for CLIP and BLIP. \\
$\bullet$ Group II (CLIP-based end-to-end networks): 
CLIP \cite{clip}, CLIP4Clip \cite{clip4clip}, X-CLIP \cite{xclip}, TS2-Net \cite{ts2net}, CLIP-VIP \cite{clip_vip_XueS0FSLL23}, DGL \cite{yang2024dgl}, and TeachCLIP \cite{2024TeachCLIP}. \\
$\bullet$ Group III (Multi-feature fusion, single space learning):  DE \cite{dualencoding} and W2VV++ \cite{w2vv++}. \\
$\bullet$ Group IV (Multi-feature fusion, multi space learning): DualTask \cite{wu2020interpretable_dual_task}, SEA \cite{sea}, LAFF \cite{laff}, and T$\times$V \cite{t_multi_v_GalanopoulosM22}. \\

\subsubsection{Results}
\cref{table:compare_baseline} presents a comprehensive comparison among various models across three benchmark datasets, IACC.3, V3C1 and V3C2, evaluated over different years from TV16 to TV23. 
We observe that models integrating multiple features generally surpass those relying on a single feature CLIP/BLIP in performance.
Even the best-performing blip zero-shot retrieval method significantly lags behind multi-feature approaches.
This trend underscores the advantage of leveraging diverse feature sets, which, in combination, offer a richer representation of video and text, thereby enhancing retrieval precision.
Particularly noteworthy is the performance of models classified under ``Learning multiple common spaces for multiple features'' which includes DualTask, LAFF, SEA, T$\times$V and our proposed \model. These models, by dedicating unique spaces for individual features, facilitate a more subtle understanding and utilization of each feature's distinct characteristics.

Comparing \model to other multi-space baselines in \cref{table:Partial_vs_Full,table:Laff_addDiverseLoss}, we observe a significant improvement. This advancement can be attributed to two key factors. 
Firstly, \model is designed to train a distinct space for each feature, complemented by a dynamic weighted fusion of features from the opposite modality.  
DualTask and LAFF manually set multiple spaces, while SEA focuses on text features, and T$\times$V performs cross-matching between text and video features. These methods do not fully leverage the potential of feature-specific space optimization.
Secondly, \model employs a \dcLoss during training to constrain the ordering of negative samples across different spaces. This enhancement foregrounds the distinctiveness of each space, further elevating the retrieval performance.

\subsection{Understanding LPD}
\subsubsection{Inter-space Divergence}

\begin{figure*}[t]
    \subfloat[LAFF, inter-space IoU: 0.487
    \label{fig:IoU_laff}]{\includegraphics[width=0.33\textwidth]{pdf/laff_topk20_tv23.pdf}}
    \hfill
    \subfloat[\model \textit{w/o} \texttt{DcL}, inter-space IoU: 0.217
    \label{fig:IoU_without_spearman}]{\includegraphics[width=0.33\textwidth]{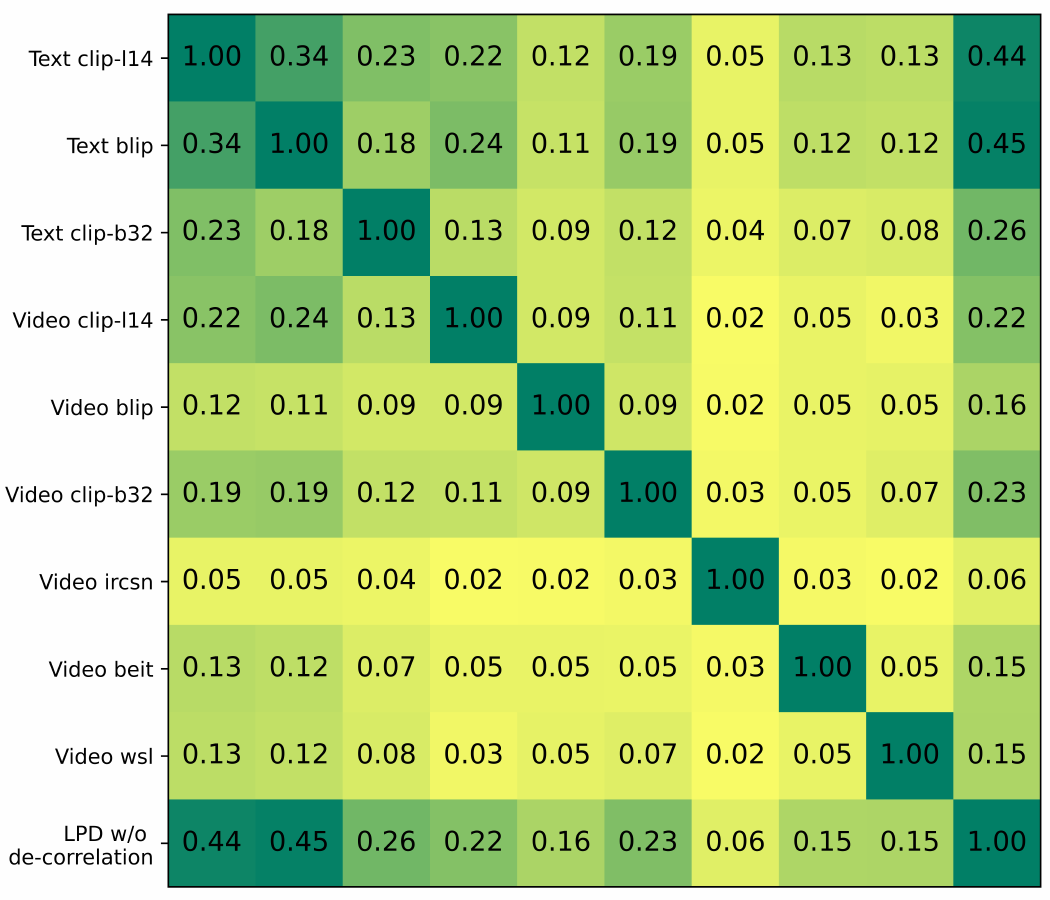}}  
    \hfill
    \subfloat[\model, inter-space IoU: 0.209
    \label{fig:IoU_model}]{\includegraphics[width=0.33\textwidth]{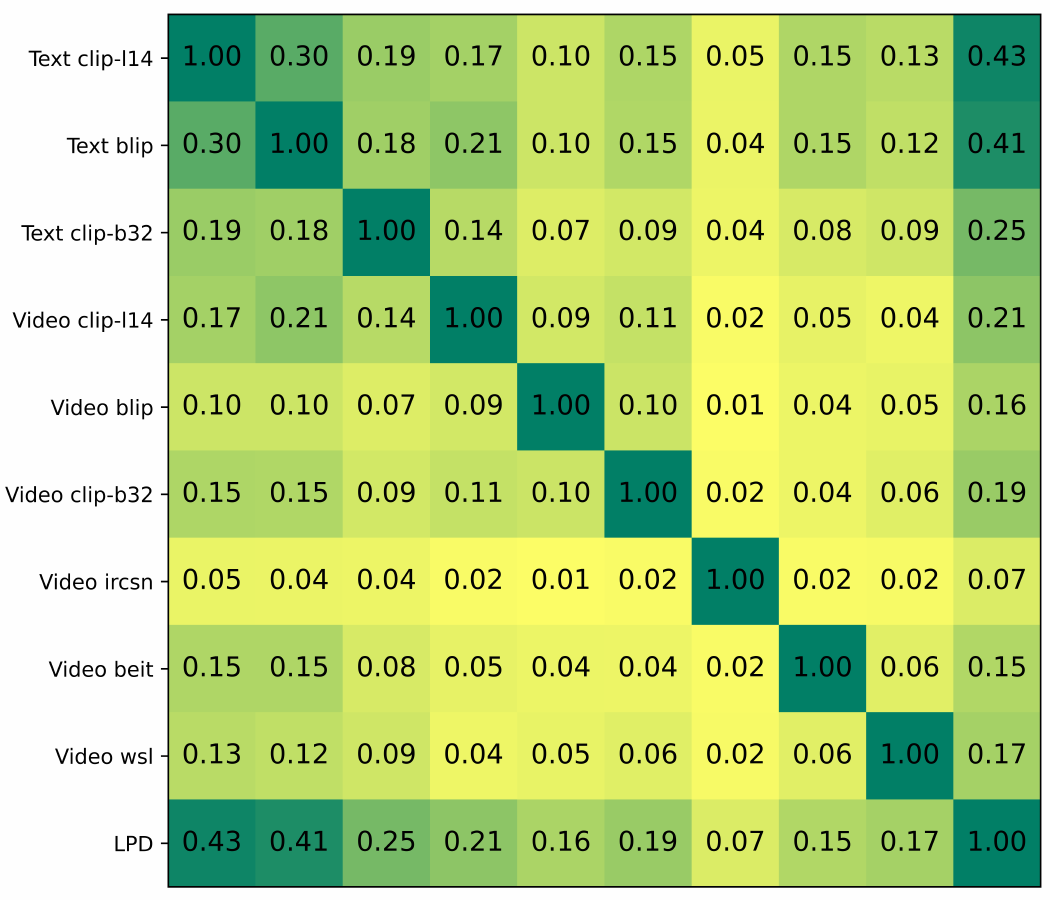}} 
    \caption{\textbf{Visualizing inter-space similarities for  LAFF, \model \textit{w/o} \texttt{DcL}, and \model}. Per model, we calculate the Intersection over Union (IoU) of the top-20 retrieval results by the individual spaces. Lower IoU scores indicate larger inter-space diversity. Best viewed in color.
    }
    \label{fig:param_compare}
\end{figure*}

In order to measure how each common space differs from another,  we compute  an inter-space similarity as follows. Given $S_i$ and $S_j$ as top-20 retrieved videos from the $i$-th space and the $j$-th space \wrt the TV23 queries respectively, the inter-space similarity is computed as the Intersection over Union (IoU) between $S_i$ and $S_j$.
The IoU heatmaps of LAFF, \model \emph{w/o} \texttt{DcL}, and \model are shown in \cref{fig:param_compare}.

\begin{table*}[!t]
\centering \setlength{\tabcolsep}{8pt}
\caption{\textbf{Ablation study.} \model (Partial De-correlation) achieves significant improvement (p$<$ 0.05).} \label{table:Partial_vs_Full}

\renewcommand{\arraystretch}{1}
\resizebox{0.85\linewidth}{!}{
\begin{tabular}{@{}p{3cm}rrrcrrrcrrr@{}}
\toprule
\multirow{2}{*}{\textbf{Setup}} & \multicolumn{3}{c}{\textbf{IACC.3}} & & \multicolumn{3}{c}{\textbf{V3C1}} & & \multicolumn{2}{c}{\textbf{V3C2}} & \multicolumn{1}{r@{}}{\multirow{2}{*}{\textbf{MEAN}}} \\
\cmidrule{2-4} \cmidrule{6-8} \cmidrule{10-11}
& \multicolumn{1}{l}{\textit{TV16}} & \multicolumn{1}{l}{\textit{TV17}} & \multicolumn{1}{l}{\textit{TV18}} & & \multicolumn{1}{l}{\textit{TV19}} & \multicolumn{1}{l}{\textit{TV20}} & \multicolumn{1}{l}{\textit{TV21}} & & \multicolumn{1}{l}{\textit{TV22}} & \multicolumn{1}{l}{\textit{TV23}} & \multicolumn{1}{l}{} \\
\midrule

\rowcolor{green!8}\model (Partial \texttt{DcL}) & \textbf{0.245} & \textbf{0.327} & \textbf{0.169} & & \textbf{0.218} & \textbf{0.287} & \textbf{0.269} & & \textbf{0.221} & \textbf{0.221} & \textbf{0.245} \\
Partial \texttt{DcL} $\rightarrow$ Full \texttt{DcL} & 0.230 & 0.309 & 0.146&  & 0.203 & 0.265 & 0.251 & & 0.201 & 0.213 & 0.227 \\
 \emph{w/o} \texttt{DcL} & 0.235 & 0.321 & 0.150 & & 0.207 & 0.262 & 0.261 & & 0.199 & 0.197 & 0.229 \\
 \emph{w/o} \efLoss & 0.244 & 0.325 & 0.165 & & 0.216 & 0.278 & \textbf{0.269} & & 0.215 & 0.220 & 0.241 \\
\bottomrule
\end{tabular}
}
\end{table*}

\begin{table*}[thb!]
\centering\setlength{\tabcolsep}{8pt}
\caption{\textbf{Performance enhancement from \dcLoss on multi-space baseline models. Adding \texttt{DcL} to individual models consistently brings in significant improvement (p$<$0.05).} }
\label{table:Laff_addDiverseLoss}

\renewcommand{\arraystretch}{1} 
\resizebox{0.85\linewidth}{!}{
\begin{tabular}{@{}p{2cm}rrrcrrrcrrp{2cm}@{}}
\toprule
\multirow{2}{*}{\textbf{Model}} & \multicolumn{3}{c}{\textbf{IACC.3}} & & \multicolumn{3}{c}{\textbf{V3C1}} & & \multicolumn{2}{c}{\textbf{V3C2}} & \multicolumn{1}{c}{\multirow{2}{*}{\textbf{MEAN}}} \\
\cmidrule{2-4} \cmidrule{6-8} \cmidrule{10-11}
& \textit{TV16} & \textit{TV17} & \textit{TV18} & & \textit{TV19} & \textit{TV20} & \textit{TV21} & & \textit{TV22} & \textit{TV23} &  \\
\midrule
LAFF & 0.222 & 0.287 & 0.142 & & 0.192 & 0.225 & 0.237 & & 0.175 & 0.183 & 0.208 \\
LAFF+\texttt{DcL} & 0.229 & 0.296 & 0.139 & & 0.198 & 0.247 & 0.230 & & 0.183 & 0.194 & 0.215 (\gain{+3.4}\%$\uparrow$) \\ [2pt]
SEA & 0.231 & 0.314 & 0.158 & & 0.199 & 0.251 & 0.230 & & 0.188 & 0.220 & 0.224 \\
SEA+\texttt{DcL} & 0.229 & 0.337 & 0.157 & & 0.200 & 0.258 & 0.249 & & 0.200 & 0.209 & 0.230 (\gain{+2.7}\%$\uparrow$) \\ [2pt]
T$\times$V & 0.240 & 0.313 & 0.159 & & 0.211 & 0.243 & 0.260 & & 0.198 & 0.200 & 0.228 \\
T$\times$V+\texttt{DcL} & 0.241 & 0.319 & 0.159 & & 0.212 & 0.259 & 0.266 & & 0.207 & 0.207 & 0.234 (\gain{+2.6}\%$\uparrow$) \\  
\bottomrule
\end{tabular}
}
\end{table*}

For the LAFF model, using a multi-head approach to construct nine parallel common spaces, results in a lack of clear distinction between each space.  As shown in \cref{fig:IoU_laff}, the high similarity between different spaces, indicated by large green areas in the heatmap, suggests that the LAFF model's spaces lack diversity.
Comparing \cref{fig:IoU_without_spearman} with \cref{fig:IoU_laff}, we observed a notable increase in the distinctiveness among the spaces, highlighted by the overall shift to yellow in the heatmap. This shift verifies that our feature-specific design successfully created diversity within the common spaces. Further, when comparing \cref{fig:IoU_model} to \cref{fig:IoU_without_spearman}, we observed a significant decrease in the IoU of various spaces (for example, the IoU between text clip-14 and text blip decreased from 0.34 to 0.30), indicating the effectiveness of our proposed \dcLoss in increasing the diversity among the common spaces. 

\subsubsection{Partial versus full \dcLoss}
As shown in \cref{fig:de_correlation_loss}, we only constrain \emph{negative} examples. Here, we provide experimental evidence to support this design choice. In \cref{table:Partial_vs_Full}, we compare different types of \texttt{DcL} settings. We observe that compared to the \model without \dcLoss, \model with full \dcLoss shows a slight decrease in performance (MEAN infAP 0.229 $\rightarrow$ 0.227), whereas \model (partial \dcLoss) improves the MEAN infAP to 0.241.
This indicates that applying \dcLoss to a full list that includes \emph{positive} examples consequently impairs retrieval performance. The reason is that applying de-correlation to the \textit{Full} ranking list would disrupt the correct ranking of the positive training samples, and consequently result in performance loss. By constraining only the correlation of \emph{negative} training examples, the proposed \dcLoss indirectly yet effectively strikes a balance between search result diversity and relevance.

\begin{figure}[t]
    \subfloat[\model \textit{w/o} \efLoss
    \label{fig:IoU_wo_EF-MTRL}]{\includegraphics[width=0.23\textwidth]{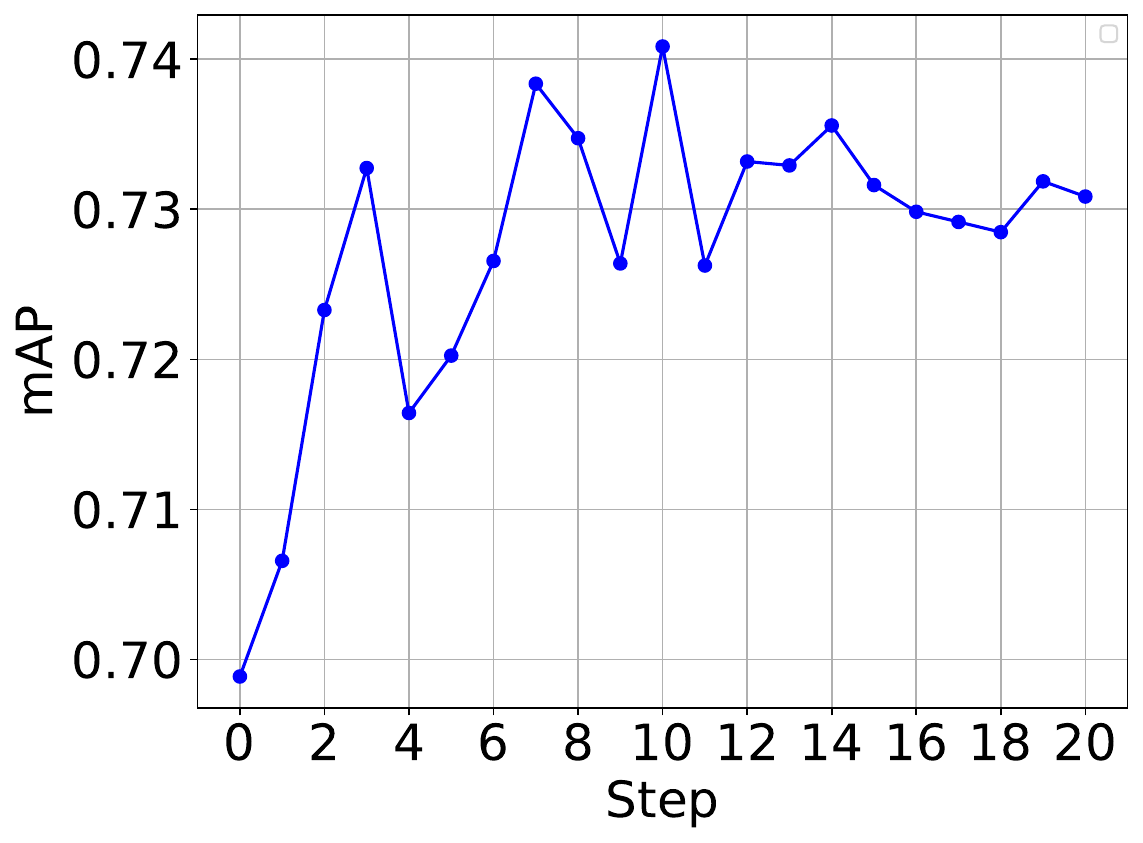}}
    \hfill
    \subfloat[\model \textit{w/} \efLoss
    \label{fig:IoU_with_EF-MTRL}]{\includegraphics[width=0.23\textwidth]{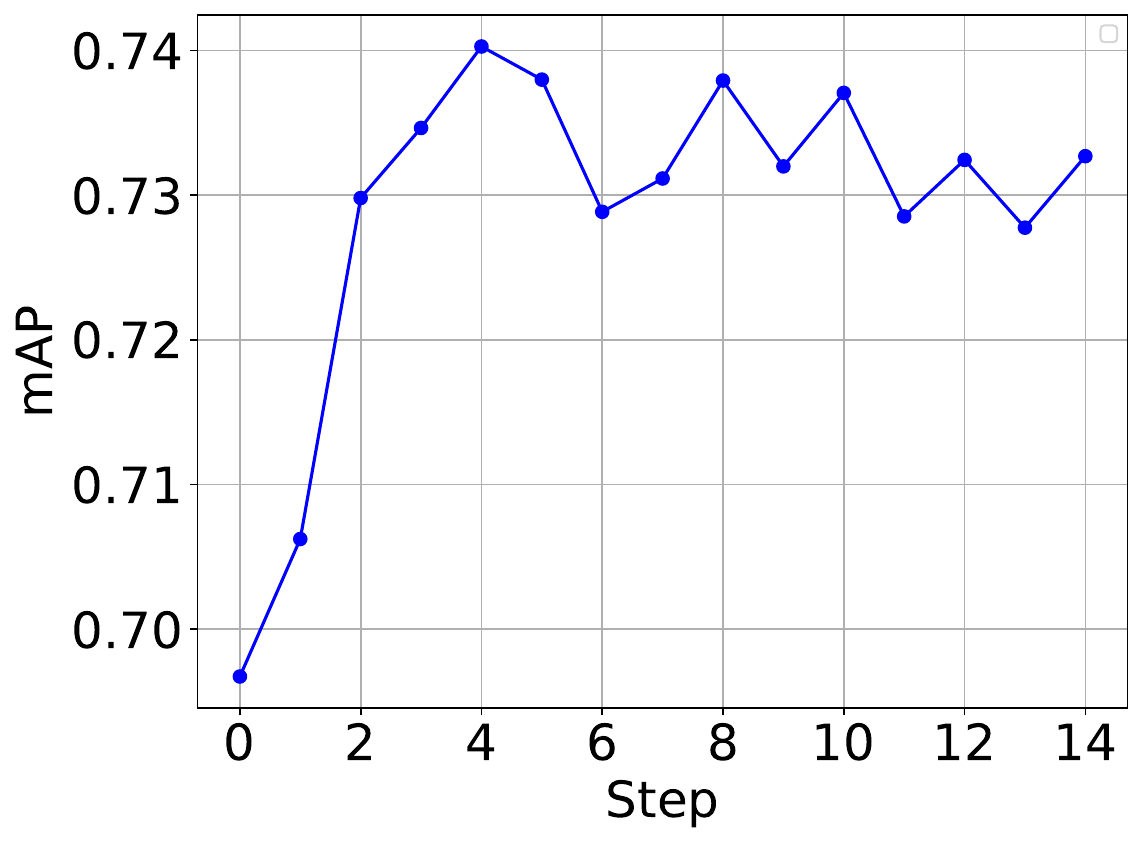}}
    \caption{\textbf{Changes in mAP on the validation set}. The \efLoss method enables faster convergence.
    } \label{fig:EF_MTRL_val_mAP}
\end{figure}

\subsubsection{\texttt{DcL} improves baseline models}

\DcLoss is designed to be model-agnostic, with its primary goal being to diversify the ordering of negative samples across different common spaces.
In \cref{table:Laff_addDiverseLoss}, we integrate \texttt{DcL} into three multi-space baselines: LAFF, SEA, and T$\times$V. The results reveal that incorporating \dcLoss into these models led to an approximate 3\% improvement in mean infAP scores. This improvement strongly indicates that \dcLoss is not only effective but also highly adaptable across different model architectures. This adaptability confirms \dcLoss as a powerful tool for increasing the diversity and accuracy of video retrieval systems in ad-hoc search tasks, demonstrating its broad applicability and effectiveness.

\subsubsection{The effect of \efLoss}
\cref{fig:EF_MTRL_val_mAP} shows the changes in mAP on the validation set, clearly indicating that the model with \efLoss converges faster and more stably compared to the one without it. As shown in \cref{table:Partial_vs_Full}, removing \efLoss (\textit{w/o} \efLoss) and using the standard triplet ranking loss causes a slight drop in infAP from 0.245 to 0.241. This suggests that enforcing fair convergence across spaces enhances model effectiveness through more balanced learning and optimization among diverse feature spaces.

We observe that different feature spaces converge at different speeds during training, with some having a much larger impact on the overall loss. Our method gradually balances these differences, leading to more synchronized convergence. This suggests that EF-MTRL acts as a regularizer by encouraging greater training stability and preventing overfitting to any single feature space.

\subsubsection{Complexity analysis}
As shown in \cref{table:Complexity_analysis}, compared to LAFF, LPD does not increase the number of parameters or FLOPs.

\begin{table}[ht!]
\centering
\setlength{\tabcolsep}{7pt}
\caption{\textbf{Complexity analysis of \textit{multi-feature, multi-space} methods.} $D$: total dimension of input features. FLOPs are estimated per video-text matching.}
\label{table:Complexity_analysis}

\renewcommand{\arraystretch}{1} 
\resizebox{0.65\linewidth}{!}{
\begin{tabular}{@{}llr@{}}
\toprule
\textbf{Method} & \textbf{Trainable Params.} & \textbf{FLOPs (M)} \\
\midrule
T$\times$V & $4\times D\times d$ & 75.50 \\
DualTask & $D\times d + 11{,}147 \times d$ & 41.70 \\
LAFF & $D\times d + 9 \times d$ & 18.89 \\
\rowcolor{green!8}LPD & $D\times d + 9 \times d$ & 18.89 \\
SEA & $D\times d$ & 18.87 \\
\bottomrule

\end{tabular}
}
\end{table}


\section{Summary and Conclusions}
We have developed \model that enhances both relevance and diversity for ad-hoc video search (AVS) at a million-video scale. \model learns diverse feature-specific common spaces and dynamically fuses information across modalities.
A \dcLoss promotes diversity in negative sample ranking, while an entropy-based multi-space triplet loss ensures relevance. Evaluations on the TRECVID AVS  benchmarks (2016–2023) show that \model outperforms a number of competitive baselines, setting a new standard for diverse and scalable video retrieval.

\begin{acks}
This work was supported by the National Natural Science Foundation of China  (No. 62172420).
\end{acks}

\onecolumn
\begin{multicols}{2}

\bibliographystyle{ACM-Reference-Format}
\bibliography{lpd}

\end{multicols}

\end{document}